\documentclass[10pt,twocolumn,letterpaper]{article}

\usepackage[pagenumbers]{cvpr} 

\usepackage{pifont}
\newcommand{\cmark}{\textcolor{green!80!black}{\ding{51}}}
\newcommand{\xmark}{\textcolor{red}{\ding{55}}}

\definecolor{cvprblue}{rgb}{0.21,0.49,0.74}
\usepackage[pagebackref,breaklinks,colorlinks,allcolors=cvprblue]{hyperref}

\usepackage{url} 

\usepackage{microtype}
\usepackage{float}
\usepackage{graphicx}
\usepackage{booktabs}
\usepackage{multirow} 
\usepackage{subcaption} 
\usepackage{color, colortbl,xcolor}
\definecolor{Gray}{gray}{0.9}
\definecolor{light-gray}{gray}{0.95}
\newcommand\sbullet[1][.5]{\mathbin{\vcenter{\hbox{\scalebox{#1}{$\bullet$}}}}}

\usepackage[mathscr]{euscript}
\usepackage{hhline} 

\def\paperID{16862} 
\def\confName{CVPR}
\def\confYear{2025}

\title{Poly-Autoregressive Prediction for Modeling Interactions}

\author{%
  \begin{tabular}[t]{@{}ccccc@{}}
    Neerja Thakkar$^{1}$ & Tara Sadjadpour$^{1}$ & Jathushan Rajasegeran$^{1}$ & Shiry Ginosar$^{2,3}$ & Jitendra Malik$^{1}$
  \end{tabular}\\[0.5em]
  { $^{1}$UC Berkeley, $^{2}$Toyota Technical Institute at Chicago, $^{3}$Google DeepMind}
}

\begin{document}

\makeatletter
\g@addto@macro\@maketitle{
    \begin{figure}[H]
    \begin{minipage}{\textwidth}
    \includegraphics[width=0.99\textwidth]{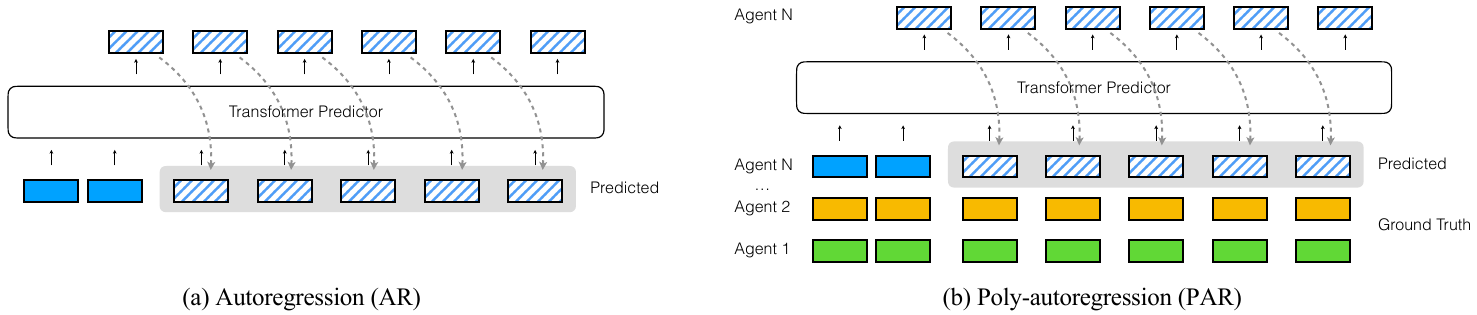}
    \centering
    \captionof{figure}{
        Inference for (a) autoregressive (AR) models and (b) our proposed poly-autoregressive (PAR) model. Solid indicates ground-truth tokens which represent a tracked data modality such as action or 6DOF pose; striped represents predicted output tokens. Color denotes agent identity. Compared to AR models, the PAR model takes other agents' tokens as inputs when making a prediction for the next timestep.
    }\label{fig:polyauto}%
    \vspace{0.2cm}
    \end{minipage}
    \end{figure}
}
\makeatother

\maketitle
\renewcommand\thefigure{\arabic{figure}}
\setcounter{figure}{1}

\begin{abstract}
We introduce a simple framework for predicting the behavior of an agent in multi-agent settings. In contrast to autoregressive (AR) tasks, such as language processing, our focus is on scenarios with multiple agents whose interactions are shaped by physical constraints and internal motivations. To this end, we propose Poly-Autoregressive (PAR) modeling, which forecasts an ego agent’s future behavior by reasoning about the ego agent’s state history and the past and current states of other interacting agents. At its core, PAR represents the behavior of all agents as a sequence of tokens, each representing an agent’s state at a specific timestep. With minimal data pre-processing changes, we show that PAR can be applied to three different problems: human action forecasting in social situations, trajectory prediction for autonomous vehicles, and object pose forecasting during hand-object interaction. Using a small proof-of-concept transformer backbone, PAR outperforms AR across these three scenarios. 
\end{abstract}  
\section{Introduction}

Large language models (LLMs) have been very successful with natural language processing (NLP) tasks, which require accurate reasoning over relationships words have within a body of text. A key component of LLMs is autoregressive (AR) modeling, where each word token is predicted based on a sequence of preceding word tokens. Building on the success of AR modeling in the NLP community, this work focuses on modeling the dynamic relationships between agents and entities in everyday interactions that occur in the physical world, such as social interactions, driving, and hand-object interactions.

Unlike language which is structured through grammar and semantics, interactions in the physical world are dictated by both the laws of physics (\textit{e.g.} how a hand grasps an object) and the internal state of each agent (\textit{e.g.} the trajectory an agent chooses to move a grasped object in) , a latent variable that we know nothing about. Furthermore, as opposed to text where each word follows another unidimensionally, the states of multiple agents are changing simultaneously. For example, in social situations, the history of a single person’s past states does not alone determine the dynamics of their future states; we also need to consider the states of other agents. We argue, therefore, that AR modeling alone is insufficient. 

In this paper, we introduce poly-autoregressive (PAR) modeling, a simple unifying approach to model the influence of other agents and entities on one’s behavior. We model behavior as a temporal sequence of states and predict an ego agent’s future behavior conditioned on the history of behavior of the ego agent as well as the rest of the agents. By considering other agents’ behaviors, we demonstrate that our approach significantly improves upon the ill-posed problem of single-agent prediction in interactive settings.

The PAR framework uses a transformer-based model for next-token prediction. Transformers have shown great success in language modeling and naturally lend themselves to predicting behavior over time. In an  interaction scenario of $N$ agents, our model predicts the future behavior of an ego ($N^{\text{th}}$) agent conditioned on its past behavior and the behavior of the other $N-1$ non-ego agents (see Figure~\ref{fig:polyauto}b). 
Each prediction task may model a different behavior modality of interest, e.g. actions for social action prediction, or 6DoF pose for object forecasting in hand-object interactions. We apply PAR to three different case studies of common real-world interactions: 
 \begin{enumerate}
     \item  \textbf{Social action prediction.} We test our method on the AVA benchmark~\cite{gu2018ava} for action forecasting. By incorporating both the ego and another agent using PAR, we get a overall $+1.9$ absolute mAP gain over AR, which only models the ego agent to predict its future behavior, and an absolute $+3.5$ mAP gain on 2-person interaction classes.
     \item \textbf{Trajectory prediction for autonomous vehicles.} When forecasting future $xy$ locations of an ego vehicle, incorporating the locations of neighboring vehicles with PAR outperforms AR, which only uses the ego vehicle's preceding trajectory as input. Specifically, on the nuScenes dataset~\cite{nuscenes}, PAR outperforms AR with a relative improvement of $6.3\%$ ADE and $6.4\%$ FDE. 
     \item \textbf{6DoF object pose forecasting during hand-object interaction.} We use the DexYCB dataset~\cite{chao2021dexycb}, where we treat the object as the ego agent and the hand as the interacting agent. While PAR integrates that hand's 3D location and object pose history, AR only uses the object's pose history to predict the object's 6DoF pose. PAR outperforms AR with relative improvements of $8.9\%$ and $41.0\%$ for the rotation and translation predictions, respectively.
 \end{enumerate}

In all these settings, we find that incorporating the behavior of other agents in the scene improves predictions of the ego agent's behavior.
All of these problems are modeled via the same simple PAR framework and implemented using the same proof-of-concept 4 million parameter transformer \textit{without any modifications to the base framework or architecture}, only to data pre-processing and choice of tokenization. We also provide an example of a simple way to build on our architecture through a location positional encoding, Sec.~\ref{sec:car_trajs}
.

The primary contribution of this work is a versatile framework that can be applied to a diverse range of settings, without modifications aside from domain-specific data processing. Our results suggest that PAR provides a simple formulation that, with a more complex transformer backbone and larger datasets, could enhance prediction of diverse multi-agent interactions across various problem domains. To facilitate further exploration and development, we have released our code, which contains the building blocks to use PAR for modeling other types of multi-agent interactions.

\section{Related Work}








\textbf{Autoregressive models.} Autoregressive modeling has a rich history in information theory and deep learning, tracing back to Shannon's paper on language prediction~\citep{shannon1951prediction} and Attneave's study on visual perception~\citep{attneave1954some}. These foundational works laid the groundwork for modern applications in deep learning, including~\citep{larochelle2011neural}, which revisits neural autoregressive models and ~\citep{gregor2014deep, theis2015generative}, which explore continuous-valued modeling. ~\citep{van2016pixel} developed PixelRNN and PixelCNN to autoregressively generate images one pixel at a time using RNN and CNN architectures, respectively. 

\medskip
The development of the transformer model~\citep{vaswani2017attention} spurred progress in computer vision with the image transformer~\citep{parmar2018imagetransformer} and the vision transformer~\citep{dosovitskiy2020image}, and autoregressive models ~\cite{chen2020generative, rajasegaran2025empirical} and more notably in NLP, where the GPT family of models~\citep{radford2018improving, radford2019language, brown2020language} has demonstrated the power of large-scale unsupervised autoregressive pre-training. Recent research has focused on multimodal learning, exemplified by the Flamingo~\citep{alayrac2022flamingo} or LlaVa~\citep{liu2023improved} models, which combine vision and language processing capabilities, illustrating the versatility of autoregressive models across various domains in artificial intelligence. While these approaches operate on image patches and word tokens, we operate on symbolic representations extracted from in-the-wild videos showing natural interactions.  A recent approach~\citep{radosavovic2024humanoid} frames humanoid locomotion as an autoregressive next-token prediction task that operates on two types of continuous tokens: observations and actions. This approach projects continuous tokens to the hidden dimension and uses a shifted loss, similar to the next-timestep prediction proposed in our framework.

\medskip
\noindent
\textbf{Multi-agent regressive models.} 
Several prior works addressed modeling specific multi-agent problems via regressive models as one-off case studies. We introduce the PAR framework to unify these efforts into a single cohesive framework. Many behavior prediction works focus on two agents engaging in social interaction, whether it be dyadic communication~\citep{Ng_2022_CVPR, ng2023text2listen, ng2024audio2photoreal} or social dance~\citep{siyao2024duolando,maluleke2024synergy}. These studies primarily tackle the challenge of predicting the state of an interacting partner (Person B) based on the input from Person A's state, sometimes extending predictions into the future~\citep{guo2022multi,maluleke2024synergy}. While earlier works used architectures such as variational RNNs~\citep{baruah2020multimodal}, recent works have predominantly adopted transformer architectures for social interaction modeling~\citep{guo2022multi,Ng_2022_CVPR, 10036100, ng2023text2listen, siyao2024duolando}, with some works exploring diffusion~\citep{liang2024intergen}, or diffusion with attention~\citep{ghosh2024remos}. Our PAR framework focuses on transformer models.

\medskip
Works encompassed by the PAR framework extend beyond human social interaction. Many multi-agent human or car trajectory prediction approaches use autoregressive prediction.  For instance, MotionLM~\citep{seff2023motionlm} utilizes a transformer decoder that processes multi-agent tokens, incorporating a learned agent ID embedding. This methodology informs our approach across all our case studies. \textit{Critically, in contrast to all prior multi-agent regressive works that designed solutions to address specific applications, we demonstrate that we can unify a diverse set of multi-agent regressive problems under a single PAR framework.} See Appendix Sec.~\ref{sec:case_study_related} for case-study-specific works.

\medskip
\noindent

\begin{figure*}
\centering
    \includegraphics[width=.99\textwidth]{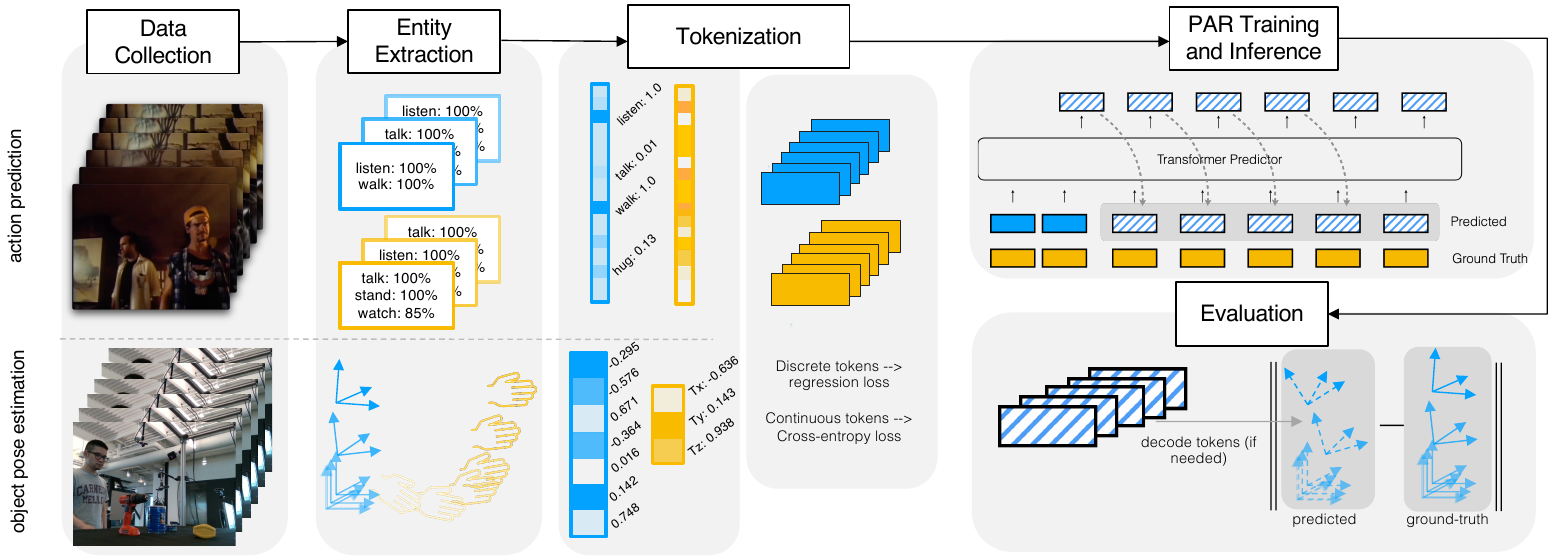}  
    \caption{\textbf{The PAR Framework}. We begin by collecting a video dataset, such as AVA (top) or DexYCB (bottom). Then, using dataset labels or computer vision techniques, a trajectory of a given modality for our prediction task is extracted for each agent, such as action class labels (top) or object pose and 3D hand translation (bottom). Data is then tokenized, either through discretization or directly using continuous values, with our framework supporting both formats. Based on the tokenization and prediction task, we choose the appropriate loss function for PAR training. After training with PAR, predicted tokens can be decoded back to data space and evaluated with relevant metrics.}
    \label{fig:PAR_pipeline}
\vspace{-1em}
\end{figure*}

\section{Poly-Autoregressive Modeling}

Our goal is to model the behavior of an agent or entity while taking into account any other agents it interacts  with, if any. To evaluate the performance of our model in capturing interaction dynamics, we predict the agent's future behavior and compare it against ground-truth data. 

We define the following task: \textit{In an interaction comprised of $N$ agents, given the observed past states of the $N-1$ interacting agents, and the observed or previously-predicted past states of the $N^{\text{th}}$ ego agent, predict the future states of the $N^{\text{th}}$  ego agent.}

We define a transformer-based poly-autoregressive (PAR) predictor, $\mathcal{P}$, that learns to model temporally long-range interactions in the input sequence. The inputs to the predictor are the past states of the $N$ interacting agents, and its output is the predicted future state of the $N^{\text{th}}$ ego agent.

\subsection{Problem Definition}
\label{sec:prob_def}

Let $\mathbf{S}=\{\mathbf{s}_i\}_{i=1}^T$ be a temporal sequence of agent states, $\mathbf{s}_i$.
We use $\mathbf{S}^N$ and $\mathbf{S}^{1:N-1}$ to denote the temporal sequences of states of the $N_{th}$ agent and of the other $N-1$ agents, respectively.
For each timestep $t \in [t_\pi,T]$, where $t_\pi \in [1,T]$ is the time we start predicting,
we take as input all other $N-1$ agents' past observed state sequences
$\mathbf{S}^{1:N-1}_{1:t-1}$
along with the $N_{th}$ agent's
past observed states up to $t_\pi$, 
$\mathbf{{S}}^N_{1:t_\pi}$,
and any of its previously predicted past states $\mathbf{\hat{S}}^N_{t_{\pi}+1:t-1}$,
if available (see Fig.~\ref{fig:polyauto}).
Our predictor, $\mathcal{P}$, then \textit{poly-autoregressively} predicts the $N_{th}$ agent's future states one time-step at a time:
\begin{equation}
       \mathbf{\hat{s}}^N_{t} = \mathcal{P}(\mathbf{S}^{1:N-1}_{1:t-1}, \mathbf{{S}}^N_{1:t_{\pi}},\mathbf{\hat{S}}^N_{t_\pi+1:t-1}). \\
\end{equation}
$\mathcal{P}$ learns to model the distribution over the next timestep of the $N_{th}$ agent's states, given all other agents' states:
\begin{equation}
p(\mathbf{\hat{s}}^N_{t} | \mathbf{S}^{1:N-1}_{1:t-1}, \mathbf{S}^N_{1:t-1}).
\end{equation}

While we provide the observed ground truth states of other agents at inference, during training, we jointly maximize the likelihood of all $N$ agents by computing losses on their future state predictions.

We train the predictor by maximizing the likelihood of the target state $y$ at time $t$:
\begin{equation*}
\label{eq:transformerloss}
    \mathscr{L_\mathcal{P}} = E_{y \sim p(y)}[-\log(p(\mathbf{s}^N_{t})],
\end{equation*}
where the target state $y$ at $t$ is computed from the $N_{th}$ agent ground truth future state.

\subsection{The Poly-Autoregressive Framework}
\label{sec:framework}

\begin{figure}
    \centering
    \begin{subfigure}[b]{0.49\textwidth}
        \includegraphics[width=\textwidth]{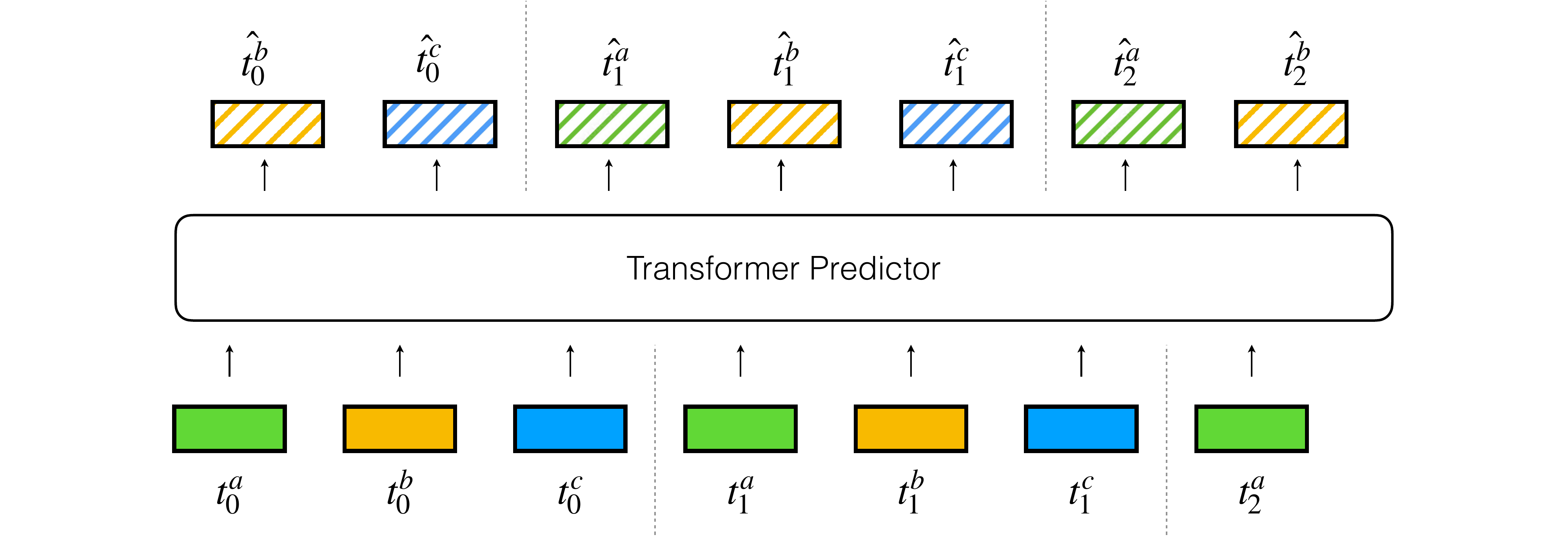}
        \caption{AR: multi-agent, next-token training.}
        \label{fig:AR_multiagent_nexttoken}
    \end{subfigure}
    ~
    \begin{subfigure}[b]{0.49\textwidth}
        \includegraphics[width=\textwidth]{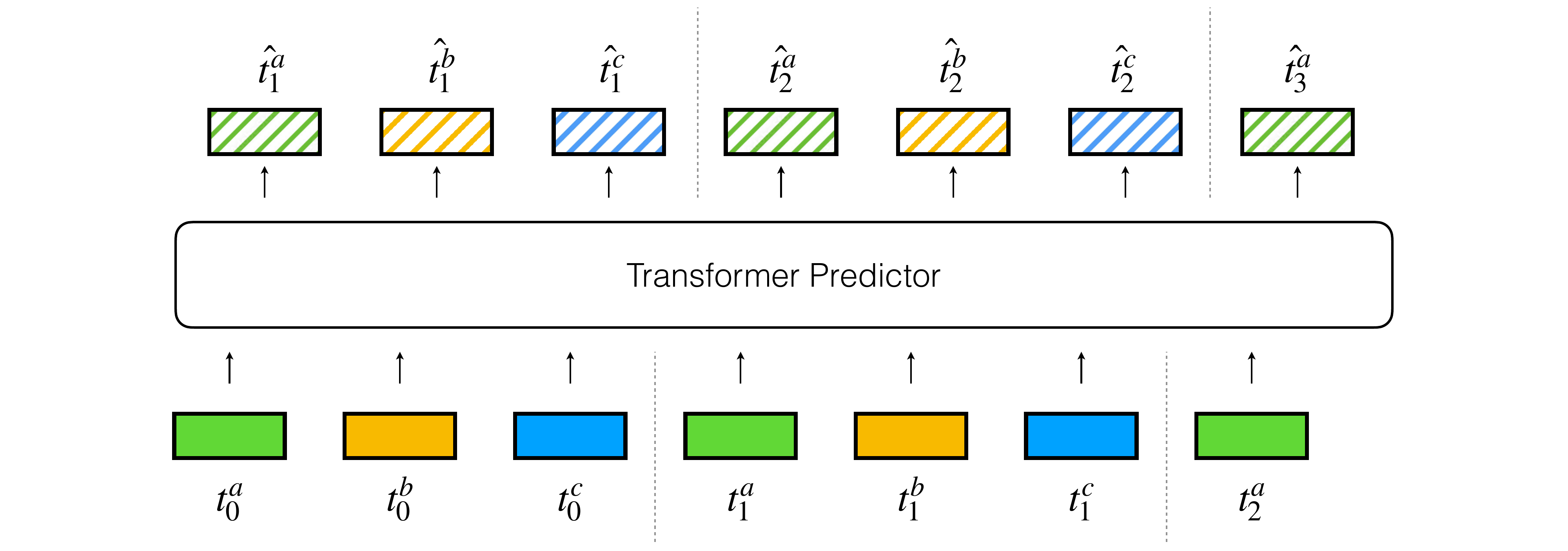}
        \caption{PAR: same-agent, next-timestep training.}
        \label{fig:PAR_nexttimestep}
    \end{subfigure}    
    \caption{Training with teacher forcing for (a) multi-agent next-token prediction in autoregressive models and (b) multi-agent poly-autoregressive models. Solid vs striped indicates a ground-truth vs predicted token, respectively. Color denotes agent identity. The AR model is trained for next-token prediction, while the PAR model is trained to predict the next timestep of the same agent. Three agents are shown for ease of visualization, but the PAR model supports an arbitrary number of agents.}
    \label{fig:PAR_training}
\vspace{-.75cm}
\end{figure}

We address the problem of forecasting the future states of an agent (from time $t$ to $T$) in a data-driven way, given a temporal sequence of past states (from time $1$ to $t-1$). 
We assume that our agent has some feature, or a set of features, of interest in a video (e.g., 3D pose) that we can tokenize. We predict the future states of the agent in terms of this tokenized feature (or set of), where we use one token (or set of tokens) per time step. The predicted tokens can be discrete (i.e., an index into a feature codebook) or continuous (i.e., a vector of one or more continuous values). The loss $\ell$ will depend on the problem's specifics and the type of token used. To train the model to predict the future, we rely on all the interaction dynamics of length $T$ in our training dataset as ground truth examples.

As a baseline, we consider the \textbf{single-agent autoregressive (AR)} paradigm, where a transformer is trained to perform next-token-prediction with teacher forcing. AR uses greedy sampling to generate sequences at inference time, predicting one next token at a time (Fig.~\ref{fig:polyauto}(a)). 

In contrast, our \textbf{multi-agent poly-autoregressive (PAR)} framework considers the other $N-1$ agents in the scene when predicting the future state of the $Nth$ agent. In this setup, we tokenize the features of interest of all $N$ agents, yielding $N$ tokens at each timestep for a total of $N*T$ tokens. In practice, we operate on a flattened sequence of $N*T$ tokens. 
Instead of using the AR training procedure in this multi-agent case (as in Fig.~\ref{fig:AR_multiagent_nexttoken}), we jointly model the $N$ agents at each timestep by introducing the following features to our PAR framework.

\vspace{0.2cm}
\noindent \textbf{Next-timestep prediction.} 
A standard AR model predicts the next token. Given the flattened sequence of $N*T$ tokens our model operates on, next token prediction would take as input an agent $k$ at timestep $t$ and predict agent $k+1$'s state at the same timestep $t$ (as in Fig.~\ref{fig:AR_multiagent_nexttoken}). However, our goal is to predict the input agent $k$'s future state at time $t+1$. Therefore, we perform \textit{same-agent next-timestep} prediction rather than next-token prediction (see Fig.~\ref{fig:PAR_nexttimestep} for an illustration of same-agent next-timestep at training).

\medskip \noindent \textbf{Learned agent identity embedding.} When giving a model information corresponding to multiple agents, the model can benefit from knowing which token corresponds to which agent. We give the model this information with a learned agent ID embedding.


\medskip \noindent \textbf{Joint training.} We train the model to jointly predict the future of all agents by computing a loss on the predicted tokens of all agents (Fig.~\ref{fig:PAR_nexttimestep}). Please refer to Section~\ref{sec:prob_def} for our inference paradigm.

\subsection{Task-Specific Considerations}
\label{sec:task-specific}
Our simple PAR approach unifies diverse problems under a single framework and architecture without any modifications. In order to formulate a problem as interaction-conditioned prediction, users must consider several task-specific details. Fig.~\ref{fig:PAR_pipeline} gives an overview of how the PAR framework disentangles multi-agent learning from problem-specific modeling.

\medskip \noindent \textbf{Data.} The input data source in our example tasks is always a collection of videos. From these videos, we extract various modalities relevant to the task at hand. These modalities can range from high-level features, such as action class labels, to low-level ones, such as 3D pose (Fig.~\ref{fig:PAR_pipeline} first two columns). We assume that each agent in the dataset is detected at each frame and is associated with an agent ID.
 
\medskip \noindent \textbf{Tokenization.} Our framework supports both discrete, quantized tokens and continuous vector tokens. The choice between discrete and continuous depends on the nature of the task.  
In the case of discrete tokens, we use a standard embedding layer to project to the hidden dimension. For continuous tokens, we train a projection layer to project the token into the hidden dimension of the transformer. For instance, if our continuous token is a 3D vector with an $(x,y,z)$ 3D location coordinate and our hidden dimension is $128$, our projection layer will project from $3$ to $128$ dimensions. We also train an un-projection layer that reverts the hidden dimension to the original token dimension.

\medskip \noindent \textbf{Loss.} The type of token and task-specific considerations dictate the loss function $\ell$ applied during model training. For discrete tokens, a classification loss is appropriate. For continuous tokens, we use a regression loss on the original token dimension. 

\medskip \noindent \textbf{Baselines.}
We compare to the following baselines, where applicable on a case-by-case basis:

\noindent $\sbullet$ \textit{Random token}: pick random tokens from the best available token space and use as the prediction. 

\noindent $\sbullet$ \textit{Random trajectory}: pick at random a trajectory from the training dataset to use as the prediction. 

\noindent $\sbullet$ \textit{NN}: Given an input agent $A$'s trajectory history, find the closest trajectory to it in the training set, belonging to $A^T$. Use $A^T$'s future as the predicted future.

\noindent $\sbullet$ \textit{Multiagent NN}: In a dataset with two interacting partners $A$ and $B$, where $B$ is the ego agent, given an input agent $A$'s trajectory history, find the closest trajectory to it in the training set, belonging to $A^T$. Use $A^T$'s interaction partner's $B^T$'s future as the prediction.


\smallskip
\noindent $\sbullet$ \textit{Mirror}: In a dataset with two interacting partners $A$ and $B$, use the ground truth future of agent $B$ as the predicted future for agent $A$.

\subsection{Framework Implementation Details}
We keep the following implementation details constant for all case studies (see also Sec.~\ref{sec:appendix_impl_details}).

\medskip
\noindent \textbf{Learned agent ID embedding.} Our learned agent ID embedding consists of the integer agent ID mapped to a hidden dim-sized vector, and summed to the token embedding.

\medskip \noindent \textbf{Architecture.} For all case studies, we use the  Llama~\citep{touvron2023llamaopenefficientfoundation} transformer decoder architecture with $8$ layers, $8$ attention heads, and a hidden and intermediate dimension of $128$. The decoder has $\sim$4.4M learned parameters, not including learned embedding layers which add a few thousand more parameters. A rotary positional encoding~\citep{su2024roformer} is used in addition to other summed encodings (i.e. agent ID embedding, locational positional encoding in Sec.~\ref{sec:car_trajs}). We train using teacher forcing. The only hyperparameter that changes between case studies is the learning rate.

\section{Case Study 1: Social Action Forecasting}

\begin{figure*}[t]
    \centering
    \includegraphics[width=.9\textwidth]{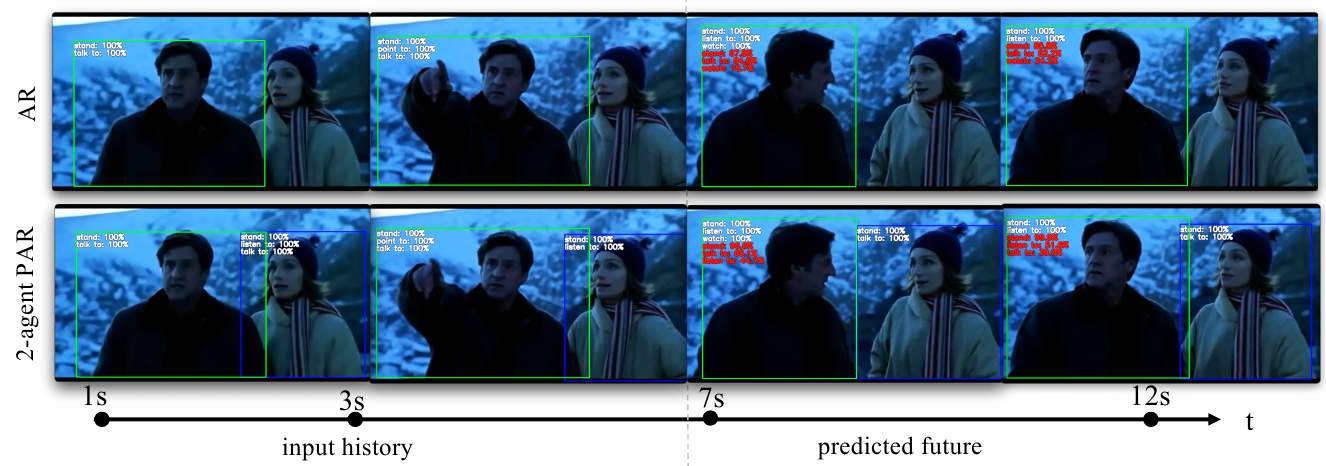}
    \caption{\textbf{Action forecasting example.} The distribution over ground truth actions are in white, and our predictions in red. A 6s action history (1Hz) is input, and 6s of future actions predicted. In the scene, the man and woman alternate between talking and listening. Initially, the man is talking. The AR model predicts he will continue talking, while the 2-agent PAR model recognizes the woman is talking and predicts more accurate turn-taking behavior.}
    \label{fig:ava_qual}
    \vspace{-.6cm}
\end{figure*}

\begin{figure}
    \centering
    \includegraphics[width=\linewidth]{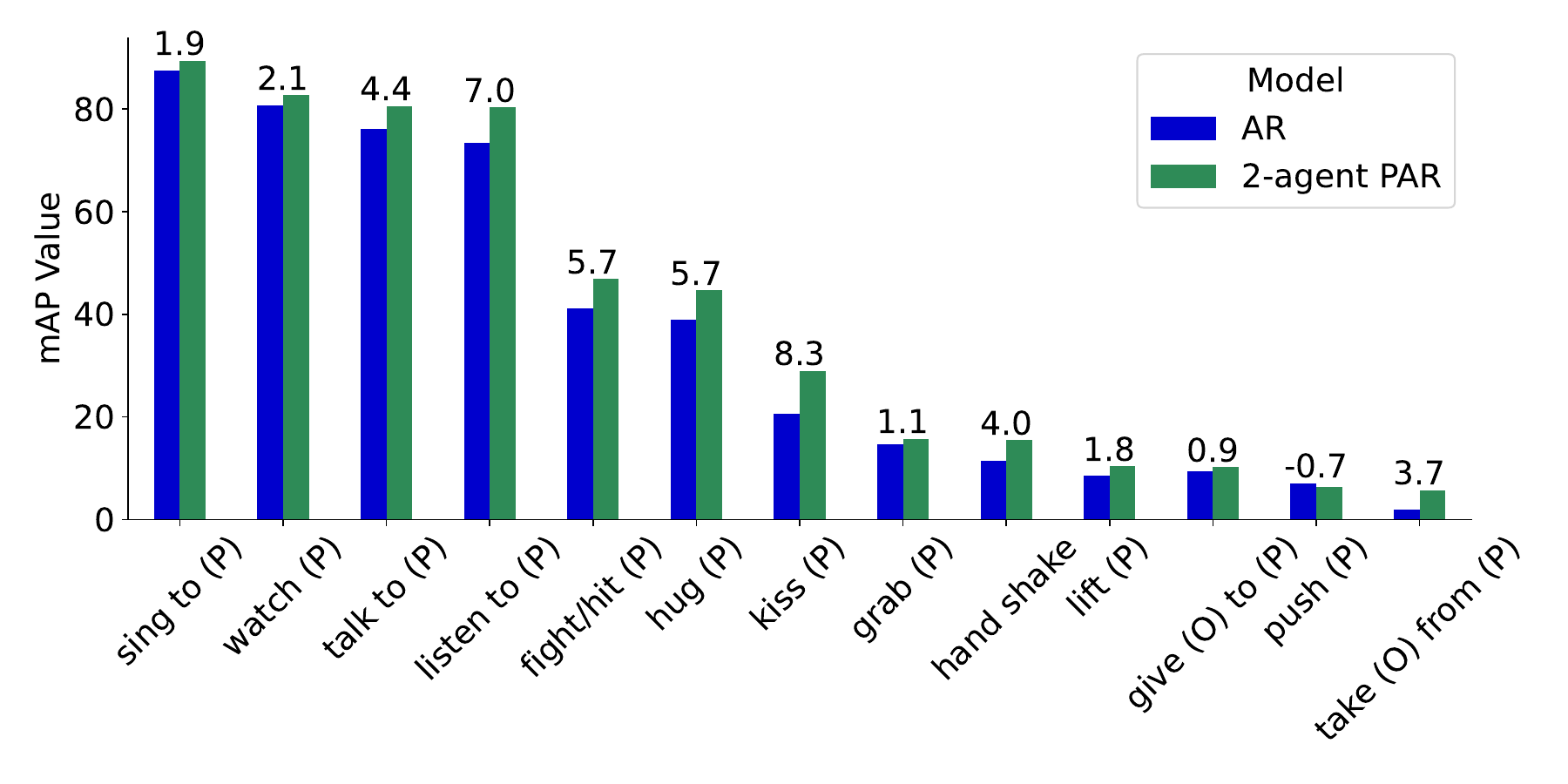}
    \small
    \caption{\textbf{Per-class mAP for AVA 2-person actions}. We see performance improvement on almost all 2-person AVA action classes ((P) stands for ``a person"). Some absolute mAP gains are particularly significant: \textit{listen to} $+7.0$, \textit{kiss} $+8.3$, \textit{fight/hit} $+5.7$, \textit{talk to} $+4.4$, \textit{hug} $+5.7$, and
    \textit{hand shake} $+4.0$.}
\label{fig:ava_2_person_classes}
\vspace{-.5cm}
\end{figure}

Our first case study involves forecasting human actions. Human behaviors are fundamentally social; for instance, individuals frequently walk in groups and alternate between speaking and listening roles when conversing. Certain actions, like hugging or handshaking, are intrinsically multi-person. Therefore, modeling human interactions should help improve action forecasting performance, especially on multi-person actions, which we show in this case study.

\subsection{Experimental Setup}
\noindent \textbf{Dataset.}  The Atomic Visual Actions (AVA) dataset~\citep{gu2018ava} comprises 235 training and 64 15-minute validation videos from movies. Annotations are provided at a 1Hz frequency, detailing bounding boxes and tracks for individuals within the frame, and each person's actions within a 1-second timeframe. Individuals may engage in multiple concurrent actions from a repertoire of 60 distinct action classes (e.g., sitting and talking simultaneously). For our analysis, we select clips featuring a continuous sequence of an agent's actions spanning at least $4$s, splitting sequences exceeding $12$s.  We use the first half of each clip as history to predict the second half. For any ego agent trajectory, we pick a second agent by selecting the person present in the scene for the longest subset of the ego agent's trajectory.

\medskip \noindent \textbf{Task-specific considerations.} Each agent's token $\mathcal{A}$ represents an 60-dimensional vector that corresponds to the actions performed at a specific timestep. Each element denotes the probability of a particular action class being enacted; ground-truth inputs are a binary vector. We implement an embedding layer that projects these tokens into the transformer's hidden dimension, as well as an un-projection layer that reverts them back to the original 60D token space for the purposes of loss calculation and output generation. We do not explicitly require the outputs to be values between 0 and 1.
We use a MSE regression loss on the 60D action tokens: $\mathscr{L} = \frac{1}{n} \sum_{i=1}^{n} (\mathcal{A}_i - \hat{\mathcal{A}}_i)^2$.
Our evaluation metric is the
mean average precision (mAP) on the 60 AVA classes.

We implement all baselines described in \ref{sec:task-specific}, where \textit{Random Token} corresponds to a random 60D vector sampled from 0 to 1. \textit{NN} and \textit{Multiagent NN} use Hamming distance as the distance metric.
\begin{table}
\centering
\begin{tabular}{@{}ccccc@{}}
\toprule
Method & Timestep pred & Ag-ID embd & mAP $\uparrow$           \\
\midrule
1-agent AR   & N/A         & N/A        & 40.7        \\
2-agent AR   & \xmark      & \xmark     & 38.0      \\
2-agent PAR*  & \xmark     & \cmark     & 40.2         \\
2-agent PAR*   & \cmark     & \xmark   & 40.0         \\
2-agent PAR &\cmark  & \cmark & \textbf{42.6}     \\
\bottomrule
\end{tabular}
\caption{\textbf{PAR action forecasting performance on AVA} We evaluate 1 and 2-agent AR methods, two 2-agent PAR ablations (rows 3 and 4, PAR*), and our PAR method. Without next-timestep prediction (see Fig.~\ref{fig:PAR_training}) or a learned agent ID embedding, our model struggles with multi-agent reasoning, performing worse than the AR baseline. With both components, the 2-agent PAR model achieves a +1.9 mAP gain over the AR method (see Fig.~\ref{fig:ava_1_person_classes} and Fig.~\ref{fig:ava_2_person_classes} for class breakdown).}
\label{tab:ava_PAR_ablation}
\vspace{-.25cm}
\end{table}
\begin{table}
\centering
\begin{tabular}{@{}lcc@{}}
\toprule
Baseline & Agents &  mAP $\uparrow$           \\
\midrule
Random Token         & 1 & 3.46          \\
Random Training Traj  & 1   & 3.44          \\
Nearest Neighbor    & 1        & 13.17 \\
Multiagent NN &2      & 5.10      \\
Mirror   & 2 & 7.97         \\
\bottomrule
\end{tabular}
\caption{\textbf{AVA baselines} While the nearest neighbor baseline performs best among baselines, it is still significantly worse than the AR model.}
\label{tab:ava_baselines}
\vspace{-.5cm}
\end{table}

\subsection{Results}

We report the performance of a single-agent AR model as a baseline, in the first line of Table~\ref{tab:ava_PAR_ablation}. The AR model is significantly better than our baselines (see Table~\ref{tab:ava_baselines}), the strongest baseline being the single-agent NN. We compare these baselines to our 2-agent PAR model (last line) and various ablations where we remove the agent ID embedding and perform next-token rather than same-agent next-timestep prediction. The second line of the table corresponds to multi-agent next-token prediction (Fig.~\ref{fig:AR_multiagent_nexttoken}). We see that this approach confuses the model, and the performance is significantly worse than just training on and considering a single agent. However, as we add various components of our PAR approach, the performance improves, and with both the next timestep prediction and agent ID embedding, we get a $+1.9$ mAP gain. When only considering 2-person action classes (enumerated in Fig.~\ref{fig:ava_2_person_classes}), our mAP is  $36.3$ on the single agent PAR model and $39.8$ on the 2-agent PAR model, a \textbf{$+3.5$} mAP gain.

In Fig.~\ref{fig:ava_qual} we see an example of action forecasting. In the input history, the man talks and the woman listens. In the future, the woman talks, and the man listens. Our 2-agent PAR model (bottom row) better understands that talking and listening actions are complementary actions, while the AR model doesn't learn this correlation. We see quantitative evidence of this in Fig.~\ref{fig:ava_2_person_classes}, with per-class mAPs for our AR vs 2-agent PAR model for 2-person action classes. Here, \textit{talk to} gets a $+4.4$ mAP gain and \textit{listen to} gets a $+7.0$ mAP gain when we train a multi-agent model. We see a significant boost on many other interaction-related action classes---for instance, \textit{kiss a person} $+8.3$ and \textit{fight/hit a person} $+5.7$ mAP---and on single-person actions, see Fig.~\ref{fig:ava_1_person_classes}.

\section{Case Study 2: Multiagent Car Trajectory Prediction}
\label{sec:car_trajs}
Our second case study focuses on predicting car trajectories. Trajectory prediction requires a vehicle to be aware of other cars on the road to avoid collisions and promote cooperative behavior. This study demonstrates how our framework enables the joint modeling of multiple vehicles' movements.

\subsection{Experimental Setup}
\noindent \textbf{Dataset.} We use nuScenes~\citep{nuscenes} , inputting 2 seconds of positions to forecast vehicle positions 6 seconds ahead. Specifically, our objective is to predict the $xy$ coordinates of each agent, exclusively considering vehicles as agents. We use the \texttt{trajdata} interface~\citep{ivanovic2023trajdata} to load and visualize the data.

\noindent \textbf{Task-specific considerations.}   Instead of discretizing the $xy$ position space, we discretize the motion, resulting in discrete velocity or acceleration tokens. These integer tokens are projected to the transformer hidden dimension using the Llama token embedding layer. Inputting only these tokens results in our PAR model knowing what speed the other agents are going at, but not where they are. It is important the model has this awareness (it should know if two agents are going to collide), so our model needs to reason over this second modality of location. We implement this by passing locations relative to the agent we are predicting into a sin-cosine positional embedding (see details in Sec.~\ref{sec:appendix_car_impl_details}), which we denote a location positional encoding (LPE). The LPE is summed to our token embeddings.

We use a cross-entropy classification loss on our discrete tokens:
$\mathscr{L} = E_{y \sim p(y)}[-\log(p(\mathbf{s}^{T}_{t_{\pi}})].$  We use the standard average displacement error (ADE) and final displacement error (FDE) to evaluate our predicted trajectories. For our baselines (Sec.~\ref{sec:task-specific}), we use the closest agent at the current timestep for \textit{Multiagent NN} and \textit{Mirror}. For \textit{NN} and \textit{Multiagent NN} we use MSE as the distance metric.

\subsection{Results}
\begin{table}
\centering
\begin{tabular}{@{}lcccc@{}}
\toprule
Token type & LPE & Method &  ADE $\downarrow$ & FDE $\downarrow$           \\
\midrule
Velocity  & \xmark & 1-agent AR & 1.50 & 3.64 \\
Velocity & \xmark & 3-agent PAR & 1.45 & 3.51 \\
Accleration  & \xmark   & 1-agent AR & 1.44    & 3.57  \\
Accleration  & \xmark & 3-agent PAR  & 1.40    & 3.44    \\
Accleration  & \cmark& 3-agent PAR   & \textbf{1.35} & \textbf{3.34} \\
\bottomrule
\end{tabular}
\caption{\textbf{Car trajectory prediction performance.} Using acceleration tokens and 3-agent PAR results in a stronger performance over velocity tokens and single-agent AR. Adding location via a positional encoding (LPE) further improves results. }
\label{tab:cars_locs}


\end{table}
\begin{table}
\centering
\begin{tabular}{@{}lccc@{}}
\toprule
Baseline & Agents &  ADE $\downarrow$ & FDE $\downarrow$           \\
\midrule
Random Trajectory & 1  & 8.89 & 16.51      \\
NN    & 1   & 1.80 & 4.13 \\
 Multiagent NN & N & 6.40 & 12.04      \\
Mirror  & N & 11.59 & 14.93      \\
\bottomrule
\end{tabular}
\caption{\textbf{Car trajectory prediction baselines.} Nearest neighbor performs best overall, but our learned single-agent AR models outperform all baselines.}
\label{tab:cars_baselines}
\vspace{-0.25cm}
\end{table}

\begin{figure}
    \centering
    \includegraphics[width=0.48\textwidth]{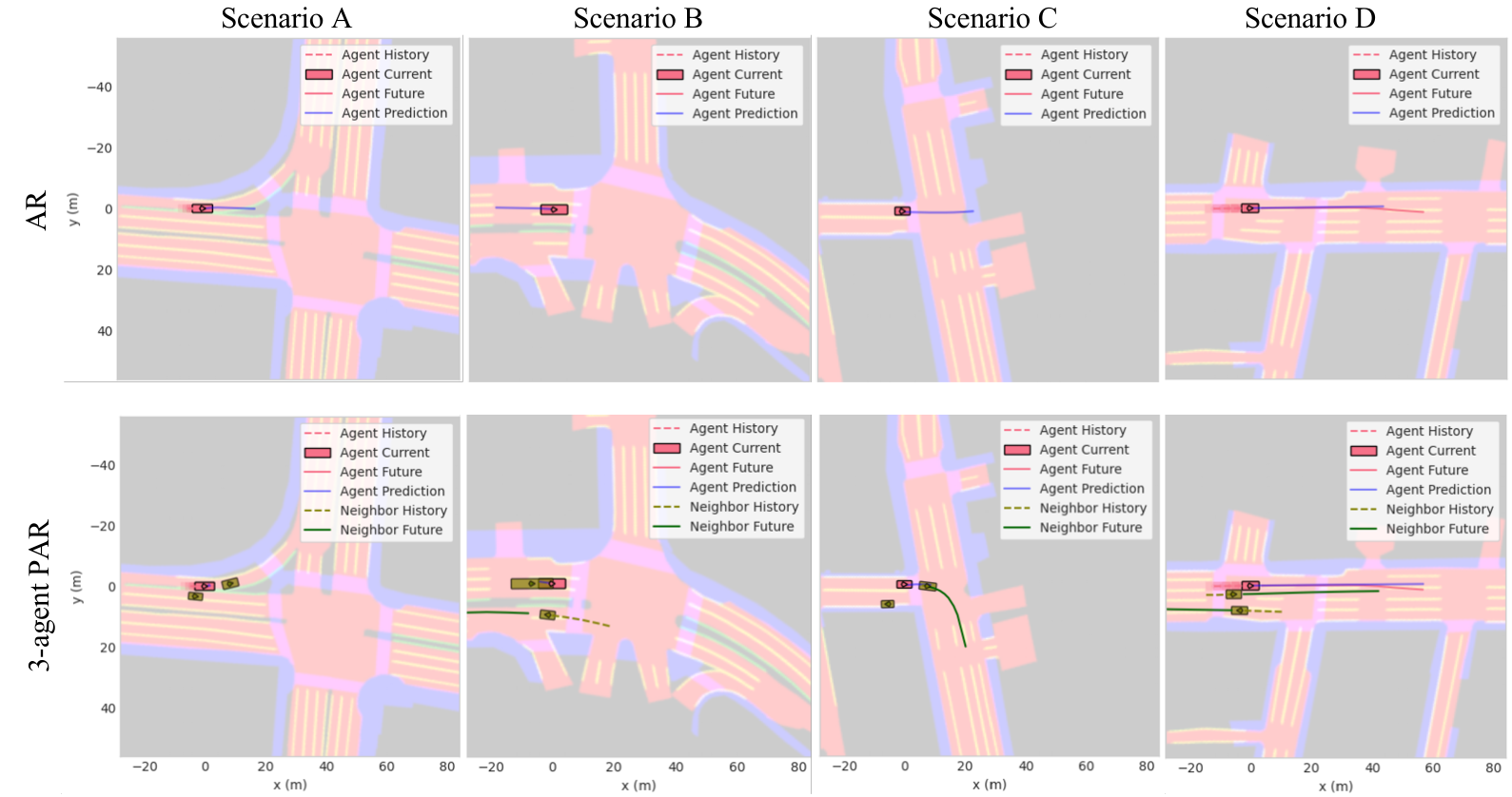}
    \caption{Example results from our single-agent AR model (top row) and three-agent PAR model with location positional encoding (bottom row) on nuScenes. The predicted agent's ground truth trajectory is in pink, and the predicted future in blue. For the PAR model, the other two agents' ground truth states are in green. Qualitatively, the PAR model handles situations where single-agent predictions might lead to collisions (A, B), uses other agents' behavior to better adhere to road areas (A, C) without environment data, and predicts based on the speed changes of other cars (D).}
    \label{fig:cars_qual}
    \vspace{-.6cm}
\end{figure}

We train AR and 3-agent PAR models using velocity tokens, acceleration tokens, and acceleration tokens combined with our location positional encoding. The results can be seen in Table~\ref{tab:cars_locs}. Note that the 3-agent PAR model uses the agent ID embedding and next timestep prediction.
Acceleration tokens consistently outperform velocity tokens both for agent AR and 3-agent PAR models. This could be because the vocabulary size for acceleration tokens is much smaller and therefore easier to optimize. Regardless, both ways of tokenizing result in models that outperform our baselines (see Table~\ref{tab:cars_baselines} - NN has a relatively low error on this dataset), and highlight that our framework is flexible such that a user can experiment with different ways of representing entities. For both token types, the 3-agent PAR model that is blind to location outperforms the AR model. While location information should help the model, it is possible that simply knowing whether other agents are slowing down or accelerating can help the model make better predictions.  When adding location information via the LPE to our 3-agent PAR model, we see another performance gain in ADE and FDE. 

Qualitative examples of the AR model (top row) and 3-agent location-aware PAR model (bottom row) can be seen in Figure~\ref{fig:cars_qual}. Our method uses no image or environment data (e.g., lanes) as input. However, by reasoning over multiple agents, its predictions lead to fewer collisions and better reasoning about speed changes and driveable areas based solely on other agents' behaviors.

\section{Case Study 3: Object Pose Forecasting During Hand-Object Interaction}

Our final case study explores how hand-object interaction can be leveraged for object pose estimation. We define the human hand and the interacting object as two agents, with tokens representing distinct state types. We show that our PAR framework can jointly model these agents, improving 3D translation and rotation predictions for the object. 

\begin{figure}
    \centering
    \includegraphics[width=\linewidth]{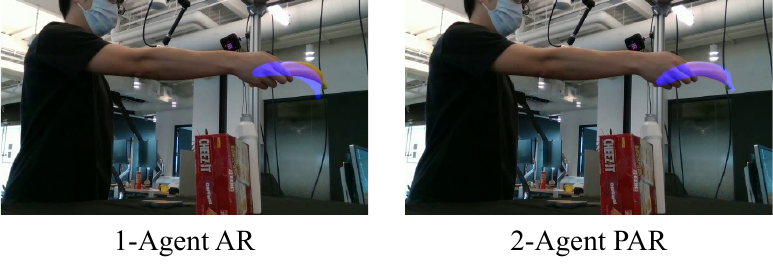}
    \vspace{-20pt}
    \caption{
    \textbf{Rotation forecasting qualitative result on test set.} 3D predictions are projected onto the image, isolating rotation results by showing the ground-truth translation. Incorporating the hand agent in the PAR framework (right) improves object pose prediction over object-only AR (left). 
    }
    \label{fig:ho_qual}
\end{figure}

\begin{figure}
    \centering
    \includegraphics[width=\linewidth]{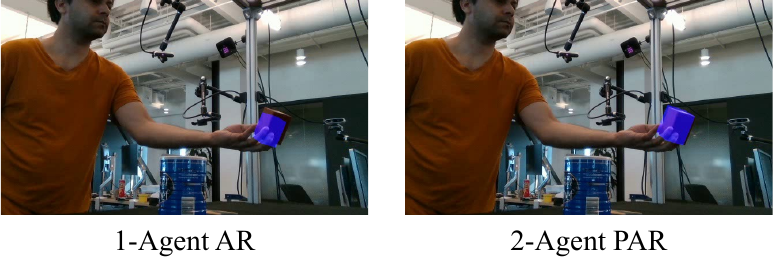}
    \vspace{-20pt}
    \caption{
    \textbf{Translation forecasting qualitative result on test set.} 3D predictions are projected onto the image, isolating translation results by showing the ground-truth rotation. Using the PAR framework (right) instead of AR (left) improves object pose prediction.
    }
    \label{fig:ho_qual2}
    \vspace{-.5cm}
\end{figure}

\subsection{Experimental Setup}
\noindent \textbf{Dataset.} We use the DexYCB dataset, which includes 1000 videos of 10 subjects performing object manipulation tasks with 20 distinct objects from the YCB-Video dataset. The data is split into 800 training, 40 validation, and 160 testing videos. We use one of 8 provided camera views. In each trial, subjects pick up and lift objects in randomized conditions. Labels include the object's SO(3) rotation and 3D translation, and the hand's 3D translation. We focus on predicting the object's rotation or translation.

\medskip \noindent \textbf{Task-specific considerations.} We tokenize object information in object-only experiments and both object and hand information in hand-object experiments. The object is represented as a 4D token for rotation forecasting (quaternion from SO(3) rotation) or a 3D token for translation forecasting (Euclidean coordinates). In hand-object experiments, the hand token is included with a 3D translation vector, and agent ID embeddings distinguish between the hand and object. 
Normalization is applied to all 3D translation vectors in both AR and PAR experiments; quaternions are normalized by definition and require no additional processing. 
An embedding layer projects the tokens into the transformer's hidden dimension, and another layer projects them back for prediction. 

For rotation-only forecasting, the loss is \(\mathscr{L}_{rot} = 1 - |\hat{q} \cdot q|\), where \(\hat{q}\) is the predicted quaternion and \(q\) the ground-truth quaternion. For translation-only forecasting, the loss \(\mathscr{L}_t\) is the mean squared error (MSE) between predicted and ground-truth translations. For PAR we predict relative object-to-hand translations at each frame, using the current hand position as origin, while for AR, we predict absolute object translations without considering the interacting agent. 
For PAR models, we add the loss \(\mathscr{L}_h\), a MSE on hand translation. The object-only AR rotation model is optimized with \(\mathscr{L}_{rot}\), while the PAR rotation model combines \(\mathscr{L}_{rot} + \mathscr{L}_h\); similarly, the object-only translation model is trained with \(\mathscr{L}_t\), and the hand-object translation model uses \(\mathscr{L}_t + \mathscr{L}_h\). 
At inference, the first half of each video is provided, and object predictions are autoregressively generated for the second half. Translation is evaluated using MSE, while rotation is measured using geodesic distance (GEO) on SO(3).

\subsection{Results}
We compare the object-only AR models to the hand-object PAR models in Table~\ref{tab:ho_res} for the two prediction tasks. We also present the baselines described in Sec.~\ref{sec:task-specific} in Table~\ref{tab:ho_baselines}. Figures~\ref{fig:ho_qual} and~\ref{fig:ho_qual2} show qualitative results on the rotation and translation predictions, respectively. In both prediction tasks, we observe that incorporating the human hand's interaction with the object enhances accuracy: for rotation, PAR results in a relative improvement of $8.9\%$ over AR, and for translation, $41\%$. See Section~\ref{sec:app_ho_qual} for additional qualitative results with more sampled frames.

\begin{table}
\begin{tabular}{@{}lccc@{}}
\toprule
Type & Method &  MSE $\downarrow$ & GEO ($rad$) $\downarrow$ \\
\midrule
Translation & 1-agent AR & 3.68 $\times$ 10$^{-3}$ & - \\
Translation & 2-agent PAR & \textbf{2.17} $\boldsymbol{\times}$ \textbf{10}$^{\boldsymbol{-3}}$ & - \\
\midrule
Rotation & 1-agent AR & - &  0.919 \\
Rotation & 2-agent PAR & - &  \textbf{0.837} \\
\bottomrule
\end{tabular}
\caption{\textbf{Test set results on DexYCB dataset.} For both rotation and translation forecasting, the 2-agent PAR model, which treats the hand as an additional agent, improves results.}
\label{tab:ho_res}
\end{table}


\begin{table}
\centering \footnotesize
\begin{tabular}{@{}lcc@{}}
\toprule
Baseline &  Translation - MSE ($m^2$) $\downarrow$ & Rotation - GEO ($rad$) $\downarrow$           \\
\midrule
Random  & 0.244 & 2.196 \\
Random Trajectory & 1.60$\times$ 10$^{-2}$ & 2.146 \\
NN  & 1.69$\times$ 10$^{-2}$ &  2.179 \\
Multiagent NN  & 1.71$\times$ 10$^{-2}$ & 2.170 \\
Mirror  & 1.20$\times$ 10$^{-2}$  & -  \\
\bottomrule
\end{tabular}
\caption{\textbf{Test set results for DexYCB baselines.} We cannot provide rotation results for the Mirror baseline, because the ground-truth does not include hand rotation, only 3D translation.}
\label{tab:ho_baselines}
\end{table}

\section{Discussion}

This work introduced the Poly-Autoregressive (PAR) framework, a unifying approach to prediction for multi-agent interactions. By applying the same transformer architecture and hyperparameters across diverse tasks, including action forecasting in social settings, trajectory prediction for autonomous vehicles, and object pose forecasting during hand-object interaction, we have demonstrated the versatility and robustness of our framework. 

Our findings underscore the crucial importance of considering the influence of multiple agents in a scene for prediction tasks. By modeling interactions, we significantly improved prediction accuracy over single-agent approaches on all three problems we considered. While we achieved promising results with a simple architecture, we have only provided a starting point that can be built upon extensively. For instance, incorporating environmental context or tokenizing pixel patches, especially as a way to relax our assumption on high-quality tracking,  are avenues for further research using PAR. It would be interesting to experiment with scaling the data and model. We do not specifically consider the relative importance of neighboring agents, which is an interesting future research direction (ex. dynamic attention mechanism).

Our PAR framework provides a simple and generalizable foundation of universal building blocks, ready for extension or refinement in future tasks. The PAR framework holds potential for advancing AI systems, enhancing prediction capabilities and enabling more accurate navigation and operation in real-world, multi-agent interactions.

\section{Acknowledgements} We thank Jane Wu, Himanshu Singh, Georgios Pavlakos, and João Carreira for useful discussions and feedback. This work was supported by ONR
MURI N00014-21-1-2801 and NSF Graduate Fellowships to NT and TS.

{
    \small
    \bibliographystyle{ieeenat_fullname}
    \bibliography{main}
}

\clearpage
\setcounter{page}{1}
\maketitleappendix

\section{Related Work: Case Studies}
\label{sec:case_study_related}
\textbf{Action recognition/forecasting.} Recent advancements in action recognition have significantly improved our ability to understand and classify human activities in videos, starting with the SlowFast network~\citep{feichtenhofer2019slowfast}, which introduced a two-pathway approach that processes visual information at different frame rates to capture slow and fast motion patterns. This resembles ventral and dorsal pathways of human brain for action understanding and object recognition, respectively. Following the rise of vision transformers~\citep{dosovitskiy2020image}, MViT~\citep{fan2021multiscale} showed promising results on action understanding benchmarks with multi-scale transformers. Recently, Hiera~\citep{ryali2023hiera}, presented a hierarchical vision transformer that leverages multi-scale feature learning to enhance action recognition performance, by utlizing masked image pretraining as in MAE~\cite{he2022masked}. LART~\citep{rajasegaran2023benefits},  expanded on these prior works by incorporating 3D human pose trajectories to achieve better action prediction performance. Some works forecast and anticipate actions~\cite{lai2024human}. ~\citep{sun2019relational} perform action forecasting on videos using relational information. ~\citep{loh2022long} train an RNN on long-form videos to contextualize the long past and improve predictions of the future.

\medskip
\noindent
\textbf{Car trajectory prediction.}
In the autonomous driving literature, forecasting the future motion of cars is a popular problem~\citep{huang2022survey,cui2024survey}, facilitated by an influx of datasets in recent years~\citep{chang2019argoverse,nuscenes,sun2020scalability}. Many important approaches have focused on modeling the environment in conjunction with multiple agents~\citep{casas2018intentnet, cui2019multimodal,salzmann2020trajectron++}; our framework only focuses on multi-agent interactions.
More recent advancements have seen the rise of transformer-based methods in trajectory prediction~\citep{ngiam2021scene,yuan2021agentformer}.
In particular, MotionLM~\citep{seff2023motionlm} forecasts multi-agent trajectories by encoding motion in discrete acceleration tokens and passing these tokens through a transformer decoder that cross-attends to the Wayformer~\citep{nayakanti2023wayformer} scene encoder. When applying the PAR framework to trajectory prediction, we use acceleration tokens to discretize car motion.

\medskip
\noindent
\textbf{6D pose estimation and hand-object interaction.}
6D pose estimation from monocular camera images has been extensively studied~\citep{xiang2017posecnn, li2018deepim, trabelsi2021pose, wang2021gdr}. Additionally, a related area of research known as 6D object pose tracking leverages temporal cues to improve the accuracy of 6D pose estimation in video sequences~\citep{wen2020se, deng2021poserbpf, wen2023bundlesdf, wen2024foundationpose}. There is also significant interest in learning state and action information of hands and objects through hand-object interaction data, sourced from both curated and in-the-wild video data~\citep{wu2024reconstructing}. Of particular relevance to 6D pose estimation is the DexYCB dataset~\citep{chao2021dexycb}, which contains 1000 videos of human subjects interacting with 20 objects on a table with randomized tabletop arrangements and 6D object poses. For the third case study in this paper, we propose using the PAR framework to model hand-object interactions, demonstrating that incorporating the hand as an agent provides a useful prior for enhancing object rotation and translation predictions compared to AR modeling.

\section{Additional PAR Framework Details}
\label{sec:appendix_impl_details}

\begin{figure*}
    \centering
    \includegraphics[width=.95\textwidth]{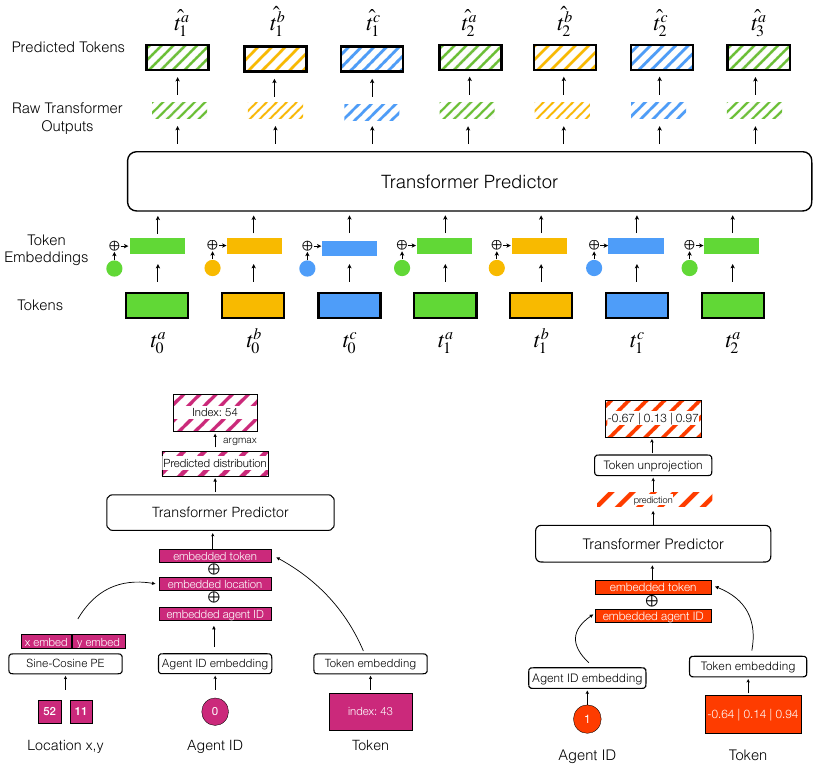}
    \caption{
    \textbf{Architecture} Top: \textit{PAR training with teacher forcing}. Here, we see that the tokens are input to the model and projected to a token embedding dimension of size $d_h=128$, where embeddings such as the agent ID embedding can be summed. Then, the transformer output is sampled (discrete tokens; left) or deprojected (continous tokens; right) to produce a predicted token. At inference, the same embeddings are summed and the same conversion from transformer output to token space occurs. Bottom left: \textit{PAR on discrete tokens}. In this case, the token in question is an integer index into a codebook and the standard transformer emebedding layer is used to project to $d_h$. Here we show the agent ID embedding and our Location Positional Encoding summed to the embedded token. After being passed to the transformer, the output is a distribution of logits, from which a token can be sampled---we use argmax in our experiments. In the discrete case, the token is converted back to the actual modality via a detokenizing step. Bottom right: \textit{PAR on continuous tokens}. Here, the token is projected to $d_h$ using a learned projection layer. The output from the transformer is of size $d_h$, and we have a trained deprojection layer to project back from $d_h$ to the token dimension. We do not add any operations after the deprojection to constrain predicted token values.
    }
    \label{fig:arch}
    \vspace{-.5cm}
\end{figure*}

\subsection{Implementation details}
\label{sec:par_impl_details}
\medskip \noindent \textbf{Token embeddings and loss.} For discrete tokens, we use a standard learned embedding layer to convert the tokens to the hidden dimension $d_{h}$ of the model. To compute the loss, we use a classification loss between the predicted distribution (output logits) and the input ground truth tokens. For continuous inputs with dimension $d$, we learn a linear layer to project from $d$ to $d_h$, and a second un-projection layer to project from $d_h$ back to $d$. To compute the loss, we take the last hidden state of the model, un-project it back to $d$, and then compute a regression loss in the original token space.

\medskip \noindent \textbf{Next-timestep prediction.} 
In standard autoregressive models (such as our single-agent model in section~\ref{sec:framework}) the next token prediction objective is enforced by computing the loss on an input and predicted target that are both shifted by one. Now, we will instead shift both by $N$, so that for a given token, the model operating on our flattened sequence of $N*T$ tokens predicts a token corresponding to the next timestep but the same agent.

\medskip \noindent \textbf{Inference.}
For a single-agent model, starting with an initial sequence history of $h$ tokens, we feed these into the model to get the next token, which we then append to our sequence to form a new sequence of $h+1$ tokens. We repeat this process to generate arbitrarily long sequences.

For our multi-agent model, we start with a ground-truth history of $h$ timesteps, which corresponds to $h*N$ tokens, including the ego agent, agent $N$. Inputting this to the model results in the last output token being our ego agent at timestep $h+1$. Then, to predict the next timestep $h+2$, we concatenate to the ground truth $h*N$ tokens the ground truth of agents $1:N-1$ at timestep $h+1$ and our prediction of the ego agent at timestep $h+1$, and we repeat this process.

For a multiagent next-token prediction ablation, to predict the ego agent at timestep $h+1$, we feed in the ground truth of agents $1:N-1$ at $h+1$ to our model to predict our ego agent, agent $N$, at timestep $h+1$. We continue this process of giving our model the ground truth tokens of agents $1:N-1$ to predict agent $N$ at each timestep. This means that this ablation necessarily has more information available at inference time.

\subsection{Architecture Details}

A detailed diagram of our architecture can be seen in Fig.~\ref{fig:arch}, where we show the overall PAR method during training with teacher forcing, and in-depth details of how discrete and continuous tokens are processed.

\section{Additional Experimental Results}

\subsection{Additional Results on AVA Action Forecasting}

\begin{figure}
    \centering
    \includegraphics[width=\linewidth]{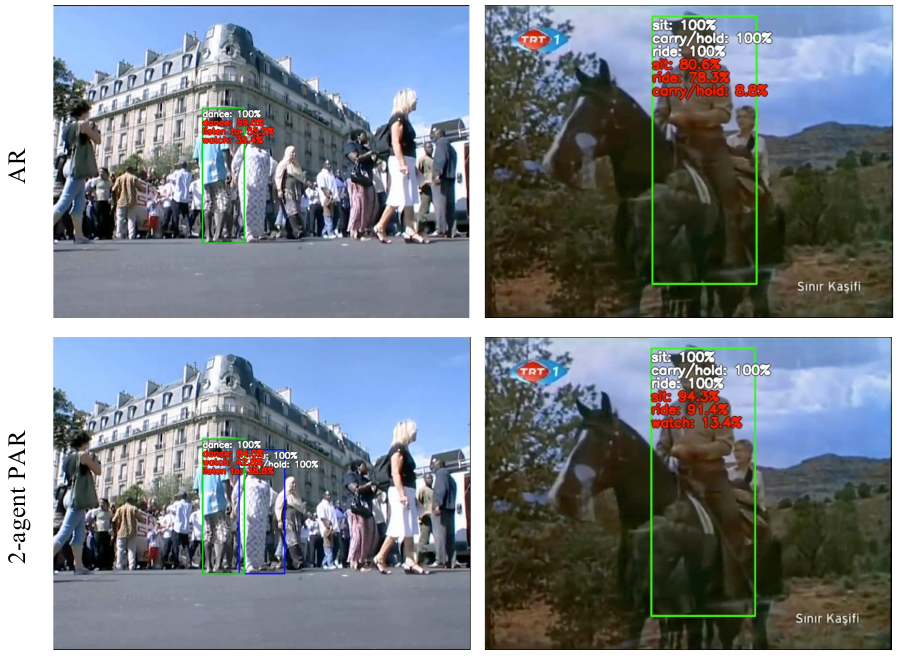}
    \caption{\textbf{Qualitative examples of action prediction on single-person actions}. While \textit{dance}, \textit{ride} and \textit{sit} are not multi-person actions, our method is able to predict them more accurately in these examples (and overall by margins of $+2.0$, $+1.0$ and $+2.7$ mAP points respectively, see Fig.~\ref{fig:ava_1_person_classes}). This is likely because these actions are co-occuring between two agents who are both partaking in the activity in question, and our 2-agent PAR model is able to reason over this. Note that in the horse-riding example, the second agent is present in the history, but the tracking failed and they are not present at the timesteps we are predicting (see full history and predictions in supplementary video). This context in the history is sufficient to help  our PAR model make better predictions.}
\label{fig:ava_qual_1_person}
\end{figure}

\begin{figure*}
    \centering
    \includegraphics[width=\textwidth]{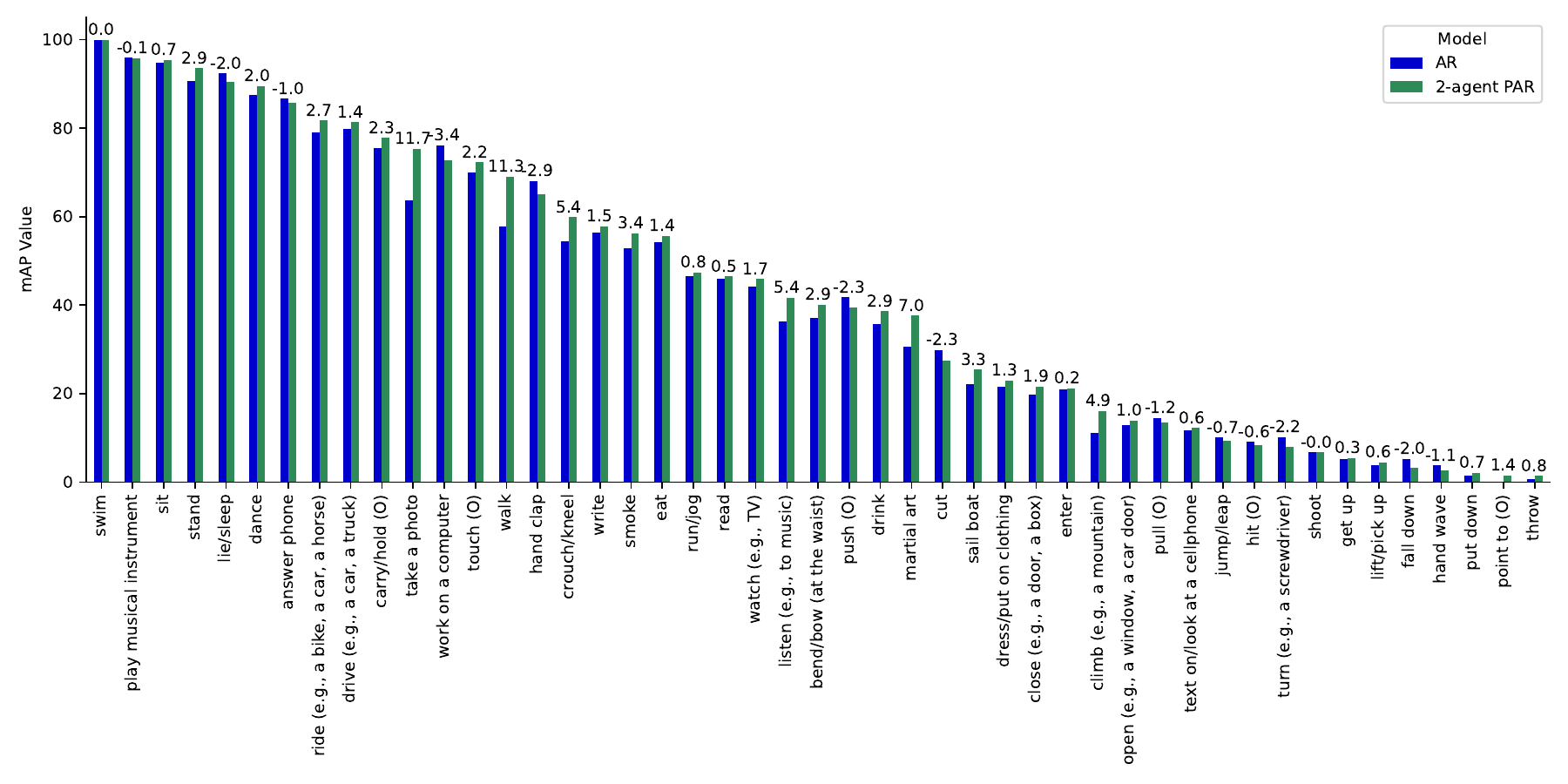}
    \caption{\textbf{Per-class mAP on AVA single-person actions}. On these actions, our PAR method is still stronger for the majority of action classes as compared to single-agent AR. For instance, we get an absolute $11.3$ mAP gain on walking - people often walk in groups, so it makes sense that this action would benefit from our PAR method. }
\label{fig:ava_1_person_classes}
\end{figure*}

We see the results of our 1-agent AR and 2-agent PAR methods on the AVA 1-person classes in Fig.~\ref{fig:ava_1_person_classes}. On the vast majority of these classes, our 2-agent PAR method is still stronger than 1-agent AR. This is likely because there are many actions that people carry out together, whether it be 2 people both \textit{dancing} (+2.0), \textit{walking} together (+11.3), \textit{watching TV} (+1.7), or \textit{listening to music} (+5.4). We show a qualitative example of some of these single-agent action classes in Fig.~\ref{fig:ava_qual_1_person}, where we see that the PAR approach helps the model make better predictions.

\medskip \noindent \textbf{Evaluation dataset} The AVA test set annotations are not released. Since we are focused on action forecasting from ground-truth past annotations instead of predicting actions from video frames, we evaluate on the validation set.

\subsection{Additional Results on Object Pose Estimation}\label{sec:app_ho_qual}

\begin{figure*}
    \centering
    \includegraphics[width=\textwidth]{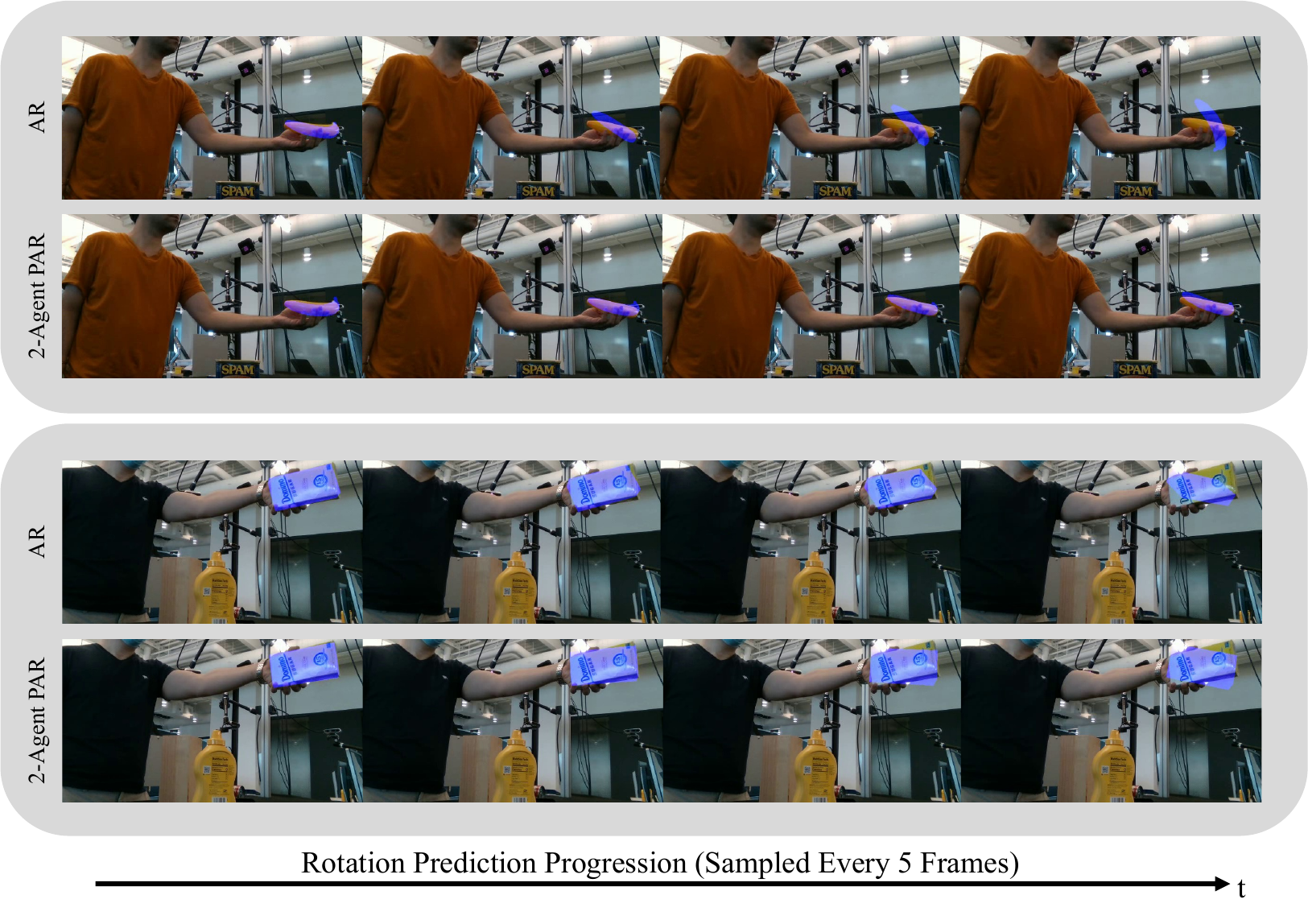}
    \caption{
    \textbf{Rotation prediction qualitative results.}
    We show results from two videos. The projected 3D model in blue has the ground-truth translation for visualization purposes and our predicted rotation. In the top row (AR), the results depict the object of interest as the sole agent, while the bottom row (2-agent PAR) demonstrates improved performance by incorporating the human hand as a second agent in the grasping interaction.}
    \label{fig:ho_qual_supp}
    \vspace{-.5cm}
\end{figure*}

\begin{figure*}
    \centering
    \includegraphics[width=\textwidth]{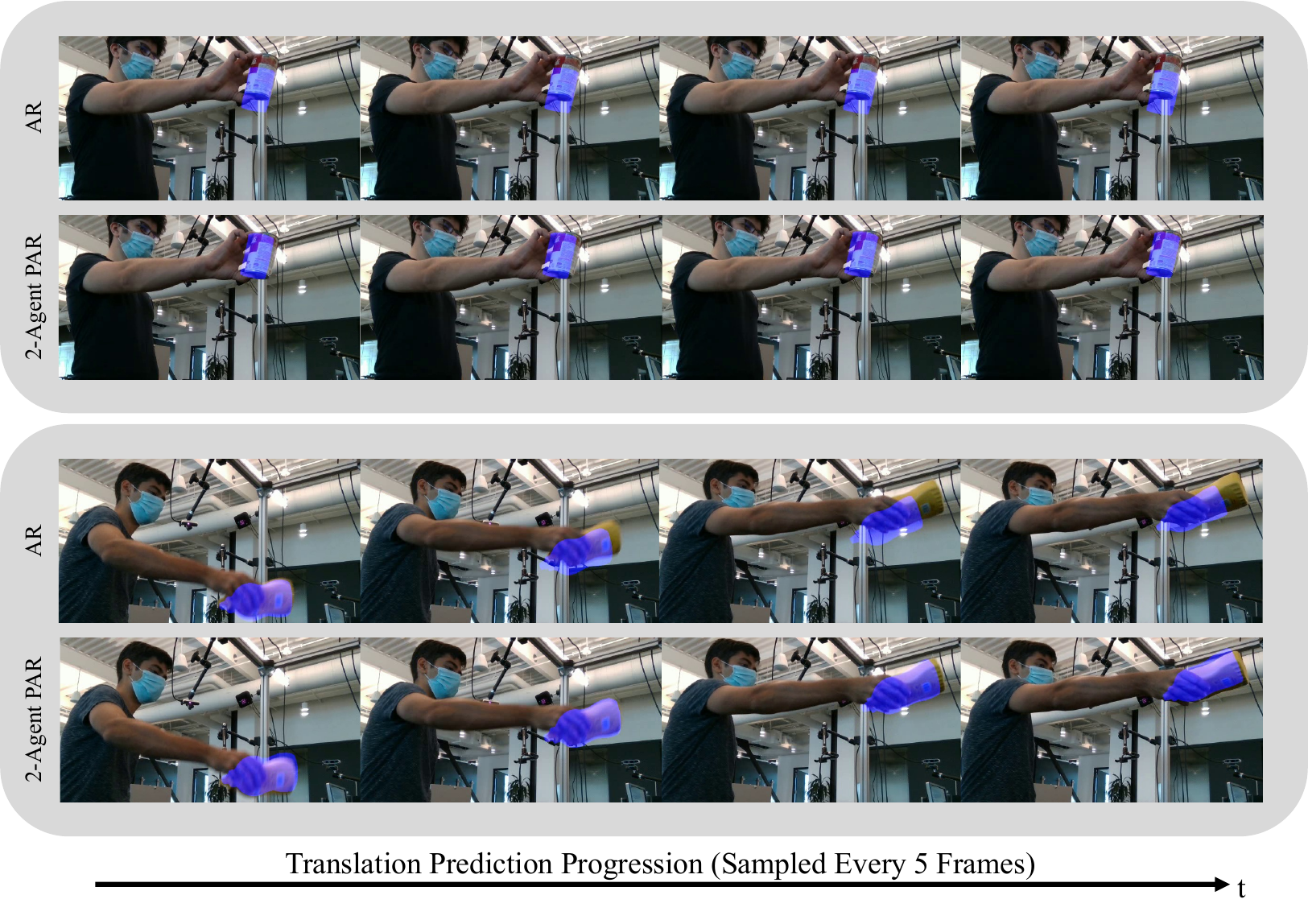}
    \caption{
    \textbf{Translation prediction qualitative result.}
    We show results from two videos. The projected 3D model in blue has the ground-truth rotation for visualization purposes and our predicted translation. In the top row (AR), the results depict the object of interest as the sole agent, while the bottom row (2-agent PAR) demonstrates improved performance by incorporating the human hand as a second agent in the grasping interaction.}
    \label{fig:ho_qual_supp2}
    \vspace{-.5cm}
\end{figure*}

See Figures~\ref{fig:ho_qual_supp} and~\ref{fig:ho_qual_supp2} for more qualitative results on rotation and translation predictions, respectively. 

Additionally, we perform an ablation on the effect of the agent ID embedding on the performance of our system for the hand-object interaction. The results are presented on the test split of the dataset, using the best checkpoint selected based on performance on the validation split after 500 epochs of training (to convergence) with agent ID embeddings. As seen in Table~\ref{tab:ho_par_ablation}, removing the agent ID embedding for the 2-agent PAR has a more significant effect on the rotation estimation than the translation estimation. This may be attributed to the fact that in the translation estimation, both the hand and the object tokens are represented as 3D translations, but in the case of the rotation estimation, the hand token is represented as a 3D translation, while the object token is represented as a quaternion. Thus, having the agent ID embedding is helpful for tokens that measure different types of quantities. 

Please note that we do not include an ablation on the next timestep prediction, because that would entail inputting the hand token and predicting the object token. Shifting by 1 instead of by 2 (our number of agents -- see Figure~\ref{fig:PAR_training} and Section~\ref{sec:par_impl_details}) would entail computing a loss on tokens of different dimensions which is not possible. The next-timestep prediction component of the PAR framework is necessary for a task such as this one with multiple data modalities.

\begin{table}
\centering
\setlength{\tabcolsep}{2pt}
\begin{tabular}{@{}lccccc@{}}
\toprule
Type & Method & Ag-ID embd & MSE ($m^2$) $\downarrow$ & GEO ($rad$) $\downarrow$           \\
\midrule
Transl  & 1-ag AR & N/A   & 3.68 $\times 10^{-3}$ & - \\
Transl   & 2-ag PAR  & \xmark & 2.26 $\times 10^{-3}$ & -        \\
Transl   & 2-ag PAR  & \cmark & \textbf{2.17} $\boldsymbol{\times 10^{-3}}$ & -        \\
Rot  & 1-ag AR & N/A & - & 0.919 \\
Rot  & 2-ag PAR & \xmark & - & 0.895 \\
Rot  & 2-ag PAR & \cmark & - & \textbf{0.837} \\
\bottomrule
\end{tabular}
\caption{\textbf{Object pose estimation PAR ablation.} All results are on the test split of the dataset. We see that for the case of object pose forecasting, using the agent ID embedding helps improve the performance on both translation and rotation prediction.}
\label{tab:ho_par_ablation}
\end{table}

\begin{table}
\centering
\begin{tabular}{@{}lccccc@{}}
\toprule
Method & Timestep pred & Ag-ID embd & ADE & FDE $\uparrow$           \\
\midrule
1-ag AR   & N/A  & N/A   & 1.44 & 3.57\\
3-ag AR   & \xmark  & \xmark & 1.36 & 3.37        \\
3-ag PAR*  & \xmark     & \cmark     &1.36 & 3.37           \\
3-ag PAR*   & \cmark     & \xmark   & 1.36 & 3.35         \\
3-ag PAR &\cmark  & \cmark & \textbf{1.35}    & \textbf{3.34}   \\
\bottomrule
\end{tabular}
\caption{\textbf{Car trajectory prediction PAR ablation.} All results use acceleration tokens, and the 3-agent methods use the location positional encoding.}
\label{tab:cars_par_ablation}
\end{table}

\setlength{\tabcolsep}{10pt}
\renewcommand{\arraystretch}{0.9} 
\begin{table}
\centering
\begin{tabular}{@{}lccc@{}}
\toprule
LPE & Method &  ADE $\downarrow$ & FDE $\downarrow$           \\
\midrule
 \xmark & 3-agent PAR  & 1.40    & 3.44    \\
 \xmark & 10-agent PAR  & 1.39    & 3.43    \\
 \cmark& 3-agent PAR   & \textbf{1.35} & \textbf{3.34} \\
\cmark& 10-agent PAR   & \textbf{1.35} & \textbf{3.35} \\
\bottomrule
\end{tabular}
\caption{\textbf{Car trajectory prediction with 3 vs 10 agents.} All results use acceleration tokens.}
\label{tab:cars_10_agent}
\end{table}

\subsection{Additional Results on Car Trajectory Prediction}

We conduct an ablation on our the agent ID embedding and next timestep prediction for the 3-agent PAR model in table~\ref{tab:cars_par_ablation}. We see that in our cars case study, our 3-agent PAR model slightly outperforms the 3-agent models without the next timestep prediction and agent ID embedding. Note that not using next timestep prediction actually results in the model having more information at inference time (see the last paragraph of Sec.~\ref{sec:par_impl_details}), so this combined with nuScenes being a relatively simple dataset, and the small acceleration-based motion token vocabulary could explain why the results are comparable.

\subsection{Results with more than Three Agents}

We experiment with using 10-agent PAR instead of 3-agent PAR for car trajectory prediction on nuScenes, Table~\ref{tab:cars_10_agent}. We see that our model has similar performance across both numbers of agents. While PAR using our small model can still learn and reason through the increased complexity of 10 agents, the increased agents do not help. We hypothesize that for cars driving on the road, there are two most influential agents: the car directly in front and the one in the adjacent lane, especially during lane changes. Therefore, we hypothesize that beyond two neighboring agents, other agents add limited value, especially on a simple dataset such as nuScenes, which is supported by these results.

Since AVA scenes often involve at most two people and DexYCB inherently includes only one hand and one object, we only go beyond two agents on nuScenes. However, PAR’s ability to handle more than 3 agents could be useful in tasks with complex group interactions, such as team sports like basketball, where many agents play key roles simultaneously.

\section{Additional Case Study Implementation Details}

We stabilize learning by using Exponential Moving Average (EMA) for training our experiments with a decay rate of $0.999$ for action prediction and object translation/rotation estimation, and $0.9999$ for car trajectory prediction.

\subsection{Car Trajectory Prediction}
\label{sec:appendix_car_impl_details}

\medskip \noindent \textbf{Tokenization} Instead of discretizing the $xy$ position space, we discretize the motion, resulting in discrete velocity or acceleration tokens computed as follows. We take each agents ground truth trajectory (past and future), shift it so that the trajectory starts at $x, y = 0, 0$, and then rotate the trajectory such that its initial heading at $t=0$ is $0$ radians. We divide velocity space into 128 even bins in $[-18, 18]$ meters. We then, separately for $x$ and $y$, take the difference between each pair of coordinates in the trajectory, to get a length $T-1$ sequence of deltas. Each of these deltas is mapped to a bin index. 

We first experimented with velocity tokens, taking the Cartesian product of bin space to give each $xy$-delta one single integer index between $1$ and $128*128 = 16384$. To get acceleration tokens, we take the difference between each $x$ delta and $y$ delta, and bin these differences into $13$ bins. We then take the Cartesian product of bin space to get a vocabulary between $1$ and $13*13=169$.

\medskip \noindent \textbf{Location Positional Encoding (LPE)} 
We implement our location positional encoding as follows. 

We first compute relative location to the agent we are predcting (the ``ego" agent) at the first timestep of the history. The ego agent trajectory is shifted to be at location $(0,0)$ at time $t=0$, and all other agents are shifted to be relative to the ego agents position. We also rotate the ego agent trajectory to have a heading of 0, and rotate all other agents trajectories relative to this ego agent trajectory. 

We normalize these relative locations (in meters) to be between 0 and 1. We then quantize these normalized locations to be an integer between 0 and 100. We next pass these locations (x and y separately) into a sin-cos positional encoding. Instead of operating on sequence position indices, the positional encoding operates on the quantized locations. We compute separate positional encodings for x and y. The encoding dimensions is half of the hidden dimension, and we concatenate the x and y encodings to get one encoding. We then sum the result of this encoding to the model inputs at training for the full trajectory (history and future).

At inference, we compute this encoding on the full trajectory (history and future) for agents 1 to N-1, but for our ego agent, we only use the history location ground truth. To get the future locations, at each sampling step, we integrate over our velocity or acceleration token to update the predicted location one step at a time, and then pass that location into our encoding.

\medskip \noindent \textbf{Evaluation dataset} Since the nuScenes test set can only be evaluated by submitting to the leaderboard, but we are interested in demonstrating the effectiveness of PAR over AR, we evaluate on the nuScenes validation set.  

\subsection{Object Pose Estimation}
As mentioned in the main text, we use relative translations with respect to the hand location for computing the PAR results. In translation prediction, the relative translation is defined with respect to the hand's position at the current timestep. This means the hand token is always treated as the origin (a zero vector) at each timestep, while the object translation corresponds to the difference between the current positions of the hand and the object.

For rotation prediction, however, we cannot compute the object rotation relative to the hand because our dataset does not provide hand rotation information. Instead, for the hand token, we compute the relative translation with respect to the hand's position in the first frame of the sequence. This ensures the hand location is only treated as the origin in the first timestep, enabling the hand's motion to influence the rotation prediction throughout the sequence.

For training both prediction tasks, we stabilize learning by employing Exponential Moving Average (EMA); rotation prediction uses a decay rate of 0.999, while translation prediction uses a decay rate of 0.99. Both prediction tasks use the AdamW optimizer with learning rate of $10^{-4}$.

\section{Supplementary Video}

Our supplementary video contains the full video for all qualitative results shown in this paper, and additional qualitative results. No temporal smoothing is applied, nor are any other modifications made to the results shown in the videos.

\end{document}